\pdfoutput=1
\documentclass[10pt]{article}
\usepackage[preprint]{tmlr}
\usepackage{amsmath,amssymb,amsfonts}
\usepackage{graphicx}
\usepackage{booktabs}
\usepackage{placeins}
\usepackage{url}
\usepackage[hidelinks]{hyperref}
\usepackage{tikz}
\usetikzlibrary{positioning}
\title{When Candidate Pools Miss Active Constraints:\\Evidence from Electromagnetic and Thermal Design}
\author{\name Yucheng Liu \email yucheng.liu@st-hildas.ox.ac.uk \\
\addr Department of Engineering Science, University of Oxford, United Kingdom
\AND
\name Chuncheng Zhao \email zhaocc@mail.iee.ac.cn \\
\addr Institute of Electrical Engineering, Chinese Academy of Sciences, Beijing, China}
\begin{document}
\maketitle
\begin{abstract}
Finite candidate pools can restrict Bayesian optimization when good designs
lie near intersecting active constraints. For independent uniform candidates
in a normalized box, the characteristic minimum sum of gaps to $k$ active
bounds scales as $(k!/N)^{1/k}$ for a pool of size $N$. We connect this
coverage limitation to acquisition search through a controlled synthetic
study and two finite-element design problems. A true-objective oracle
diagnostic over 128 seeds per configuration isolates coverage from surrogate error: boundary-enriched pools
reduce regret when several bounds are active, but do not improve interior
or single-bound cases. In matched seven-variable coil and six-variable
heat-spreader experiments, each policy receives 25 initial observations and
50 further evaluations. On ten paired seeds, continuous constrained EI
raises mean coil response from 0.9120 to 0.9745 mT and lowers mean thermal
resistance from 12.688 to 11.905 K/W relative to pool EI. Standard continuous
LogEI reaches 0.9770 mT and 11.860 K/W. An additional boundary-enriched
pool control reaches 12.092 K/W on the thermal problem. Numerical checks preserve the large
advantage over pool search while leaving the small physical differences
among continuous policies unresolved. The results identify boundary
coverage as a practical acquisition-search diagnostic and support continuous
constrained optimization, including stable LogEI, when random pools poorly
represent high-value feasible boundaries.
\end{abstract}

\section{Introduction}
Repeated geometry construction, meshing, and solution make evaluation budgets
central to simulation-driven design. Gaussian-process (GP) surrogates and
Bayesian optimization (BO) use previous observations to choose where to spend
that budget \citep{rasmussen2006gp,jones1998ego}. Their recommendations still
depend on an inner search over an acquisition function. If that search is
restricted to a random candidate pool, an accurate surrogate cannot select
a high-value design that the pool does not contain.

Intersecting active constraints make this restriction particularly costly.
When improvement requires approaching several bounds together, increasing
the number of candidates has a weak effect on coverage: under an independent
uniform-box model, the characteristic smallest sum of gaps to $k$ active
bounds decreases as $N^{-1/k}$. This calculation gives a geometric reason
to inspect acquisition search before attributing poor design quality to
the surrogate or the acquisition criterion.

We test this explanation at three levels. A synthetic family varies the
number of active box bounds. A true-objective oracle comparison separates
pool coverage from GP error and acquisition numerics, and tests an
objective-independent boundary-enrichment rule. Matched coil and heat-spreader
experiments then compare pool EI, continuous EI with finite-difference or
automatic derivatives, stable LogEI, and GP-mean search at a fixed FEM budget.
All continuous policies use the same start rule, analytic constraints, and
feasibility safeguards. The largest response gains come from replacing
finite-pool selection with continuous feasible search; LogEI gives the best
mean result among the tested EI-based policies on both production meshes.

The contribution is a diagnosis of when candidate coverage restricts the use
of expensive observations, supported by a sampling calculation and paired
experiments. Continuous acquisition optimization and LogEI are established
methods; here they provide controls and practical remedies. The oracle
experiment also shows that discrete search can benefit from explicit
boundary coverage, with a tradeoff when the optimum is interior. Extended
coil comparisons, a fixed-current variant, and numerical sensitivity checks
provide supporting engineering evidence in the appendices.

\section{Related Work}
Expected-improvement optimization of expensive black-box functions is
established \citep{jones1998ego}, as are GP regression and Latin hypercube
sampling \citep{rasmussen2006gp,mckay1979lhs}. Electromagnetic applications
combine Kriging with global optimization \citep{siah2004kriging}, examine
surrogate uncertainty \citep{hawe2007uncertainty,xiao2012exploration}, and
handle inexpensive geometric constraints explicitly
\citep{khoshoo2024electric}. FEM-based electromagnetic design has also used gradient-based coil
optimization \citep{shi1998gradient} and genetic search
\citep{petkovska2004fem}.

Gradient-based acquisition maximization is well established
\citep{wilson2018maximizing,balandat2020botorch}. LogEI addresses numerical underflow and vanishing
gradients in ordinary EI \citep{ament2023logei}. Candidate-based alternatives remain useful
when continuous optimization is inconvenient, including structured
triangulation candidates \citep{gramacy2022tricands}. Our comparison concerns
the specified random feasible pools, not every discrete or structured
candidate method. Our analysis connects the number of active bounds to candidate coverage
and matched acquisition-search experiments. Boundary enrichment serves as
a diagnostic control; we make no novelty claim for including boundary points.
Boundary over-exploration is a different concern when domain limits are
conservative and the optimum is expected to be interior; virtual derivative
sign observations can encode that expectation \citep{siivola2018boundary}.
Here the controlled study instead varies whether the optimum lies on the
boundary, exposing the tradeoff in allocating candidates there.

Unlike constrained BO with unknown feasibility \citep{gardner2014constraints},
our feasibility rules are analytic and inexpensive. They can be checked
before a simulation. The experiments use a global GP, not adaptive local
trust regions \citep{eriksson2019turbo}, and do not establish performance on
broad multimodal or simulation-constrained problem classes.

\section{Boundary coverage with finite candidate pools}
\label{sec:pool_floor}

Consider a differentiable objective on the unit box, with no additional
constraints in this calculation. Let $\mathbf u^\star$ be a maximizer with
$k\geq1$ active box bounds and
strictly positive first-order multipliers $\lambda_i$. Writing $g_i\ge0$
for normalized gaps to those bounds, the first-order active-coordinate contribution to the loss is
\begin{equation}
f(\mathbf u^\star)-f(\mathbf u)\approx\sum_{i=1}^k\lambda_i g_i.
\label{eq:first_order_loss}
\end{equation}
This expression holds with inactive coordinates fixed at their optimum;
their displacement can contribute additional, typically quadratic loss.
For an independent uniform point in the unit box and $0\le t\le1$,
$\Pr(\sum_i g_i\le t)=t^k/k!$. Consequently, if $G_N$ is the smallest
sum among $N$ independent candidates,
\begin{equation}
\Pr(G_N>t)=(1-t^k/k!)^N\approx e^{-Nt^k/k!}.
\end{equation}
A characteristic gap, at which this survival probability is about $e^{-1}$,
in the small-gap regime is therefore
\begin{equation}
t_N=\left(\frac{k!}{N}\right)^{1/k}.
\label{eq:pool_floor}
\end{equation}
This is a typical scale, not a hard lower bound: a particular pool can be
arbitrarily close to the active face. For common multiplier $\lambda$, the
corresponding first-order loss scale is $\lambda t_N$. For unequal positive multipliers, the change of variables
$v_i=\lambda_i g_i$ gives
\begin{equation}
 \Pr\!\left(\sum_i\lambda_i g_i\leq \epsilon\right)
 =\frac{\epsilon^k}{k!\prod_i\lambda_i},
 \qquad 0\leq\epsilon\leq\min_i\lambda_i.
\end{equation}
The numerator is the volume of a $k$-dimensional simplex and the denominator
includes the Jacobian of the coordinate transformation. Thus unequal
positive multipliers change the local constant while preserving the
$N^{-1/k}$ exponent. The characteristic scale is an asymptotic quantile;
it is neither the expected objective regret nor a BO convergence rate.

For $N=5000$, $t_N$ is approximately $0.0002$, $0.106$, $0.474$, and
$0.724$ for $k=1,3,5,6$. These are sums of normalized gaps, not Euclidean
distances or fractions of each coordinate range. At $k=7$ this finite-$N$ scale exceeds the $t\leq1$ interval;
the plotted single-pool reference therefore stops at $k=6$. A tenfold larger pool
reduces this scale by only 37\% at $k=5$. If $T$ independent pools were all
available for direct objective selection, the analogous scale would replace
$N$ by $TN$; an adaptive acquisition policy need not attain it.

The calculation concerns independent uniform box sampling. LHS dependence,
conditioning on feasibility, nonlinear boundaries, and resolving the
inactive coordinates prevent its use as an exact bound for the engineering
experiments. Feasibility rejection can alter the local density in either
direction. Under a regular local geometry and positive sampling density,
a similar active-gap argument motivates the exponent, but constants and
objective regret require separate analysis. Neither interior nor single-bound
optima are guaranteed easy for a finite pool. We use the formula as a geometric diagnostic and test its qualitative predictions empirically.

\section{Controlled synthetic evidence}
\subsection{Synthetic Active-Constraint Study}
\label{sec:scaling_protocol}

To test Eq.~\eqref{eq:pool_floor} with the number of active bounds as the
only varying factor, the smooth non-separable function
\begin{equation}
\begin{gathered}
f_k(\mathbf{u})=-\sum_{i=1}^{7}(u_i-c_i)^2
+0.1\sin\Bigl(\sum_{i=1}^{7}u_i\Bigr),\\
c_i=\begin{cases}1.3,& i\leq k,\\0.5,& i>k,\end{cases}
\end{gathered}
\label{eq:synthetic_function}
\end{equation}
is maximized on \([0,1]^7\). Its first \(k\) coordinates have their
unconstrained maximizer outside the cube, so exactly \(k\) upper bounds are
active at the optimum, and the remaining
coordinates are interior. The active multiplier is
$\lambda_k=0.6+0.1\cos(\sum_i u_i^\star)\in[0.5,0.7]$;
we use 0.6 as a nominal reference in the scale plot. The Hessian is
$-2I-0.1\sin(\sum_i u_i)\mathbf{1}\mathbf{1}^{\mathsf T}$,
whose largest eigenvalue is at most $-1.3$, so the constrained maximizer
is unique. For \(k=0,\ldots,7\) and eight seeds, pool search
and continuous bound-aware search start from the same 25-point LHS design and
perform 30 EI acquisitions with the GP, EI definition, pool size
\(N=5000\), and pool seed rule of the coil study. Pool sizes \(500\) and
\(50\,000\) are added at \(k=5\). Performance is the final simple regret
\(f_k(\mathbf{u}^\star)-\max_i f_k(\mathbf{u}_i)\).

\subsection{True-objective oracle and boundary enrichment}
\label{sec:oracle}
To measure candidate coverage without a surrogate, we evaluate $f_k$ at every
point in a pool and retain its true best value. Each of 128 paired seeds
(10000--10127) generates $N=5000$ candidates using independent uniform
sampling, LHS, or boundary-enriched LHS. The enriched pool keeps half of the
LHS points unchanged. For each remaining point, it chooses a number of
coordinates uniformly from $1,\ldots,7$, selects those coordinates uniformly
without replacement, and independently fixes each to 0 or 1 with equal
probability. Repeated vertices are retained and count toward the $N$
candidate slots. This rule has no knowledge of $k$, the active coordinates, or
the objective. We also test $N=500$ and $50\,000$ at $k=5$. The unique
reference optimum follows from strict concavity: its active coordinates
are one and each free coordinate $a$ solves
$a=0.5+0.05\cos(k+(7-k)a)$ on $[0.45,0.55]$.

\begin{figure}[tb]
\centering
\includegraphics[width=.78\textwidth]{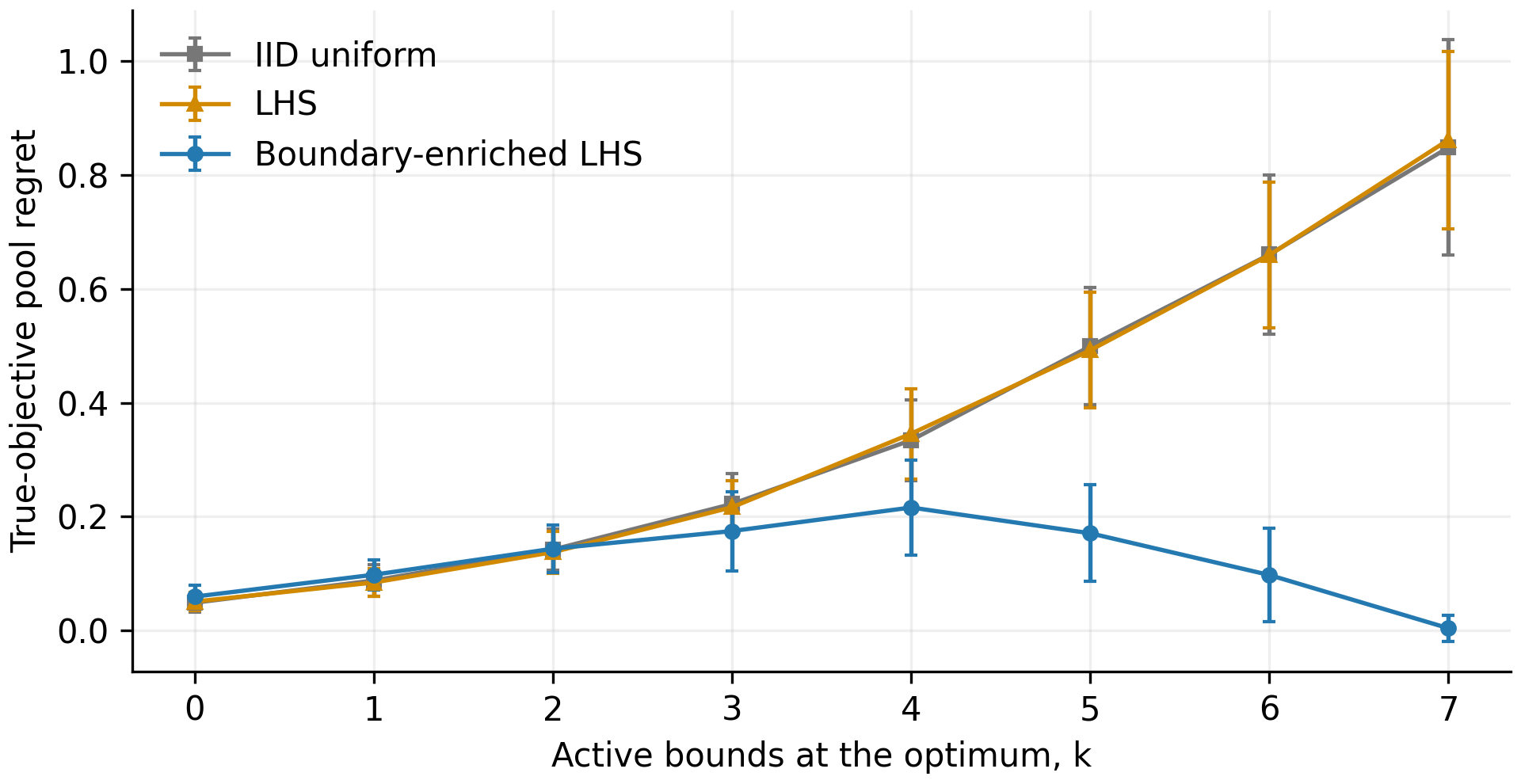}
\caption{True-objective oracle regret for 5000 candidates, mean $\pm$ sample
SD over 128 paired seeds. Boundary enrichment improves coverage for several
active bounds but does not improve the interior and single-bound cases.}
\label{fig:oracle}
\end{figure}

Figure~\ref{fig:oracle} shows mean oracle regret and sample standard deviation.
At $k=5$, independent uniform and LHS pools give $0.499\pm0.103$ and
$0.492\pm0.102$, while enrichment gives $0.171\pm0.085$. For $k=0$ and
$k=1$, enrichment instead raises the mean regret from 0.051 to 0.060 and
from 0.084 to 0.098 relative to LHS. Resolving interior coordinates remains
necessary even when a pool contains points on the correct active face.
At $k=5$, increasing $N$ from 500 to 5000 to $50\,000$ gives mean LHS
regrets of 0.764, 0.492, and 0.309; the enriched values are 0.400, 0.171,
and 0.048. The result supports a conditional coverage explanation: explicit
boundary sampling helps when several bounds are active, but spends candidate
capacity inefficiently in the interior cases tested here.

The independent-uniform active-gap check also isolates the quantity in
Eq.~\eqref{eq:pool_floor}. For $k=1,\ldots,6$, the fraction of pools with
$G_N>t_N$ ranges from 0.352 to 0.422, compared with
$(1-1/N)^N=0.36784$. At $k=7,N=5000$, $t_N>1$, outside the simplex formula's
stated interval, so we do not use that finite-$N$ check. The oracle evaluates
all $N$ candidates using a cheap analytic objective. These regrets measure
coverage at a matched candidate count and must not be compared as equal-budget
BO performance against the 55-observation sequential experiment.

\subsection{Active-Constraint Scaling}
\label{sec:scaling_results}

\begin{figure}[!tb]
    \centering
    \includegraphics[width=.65\textwidth]{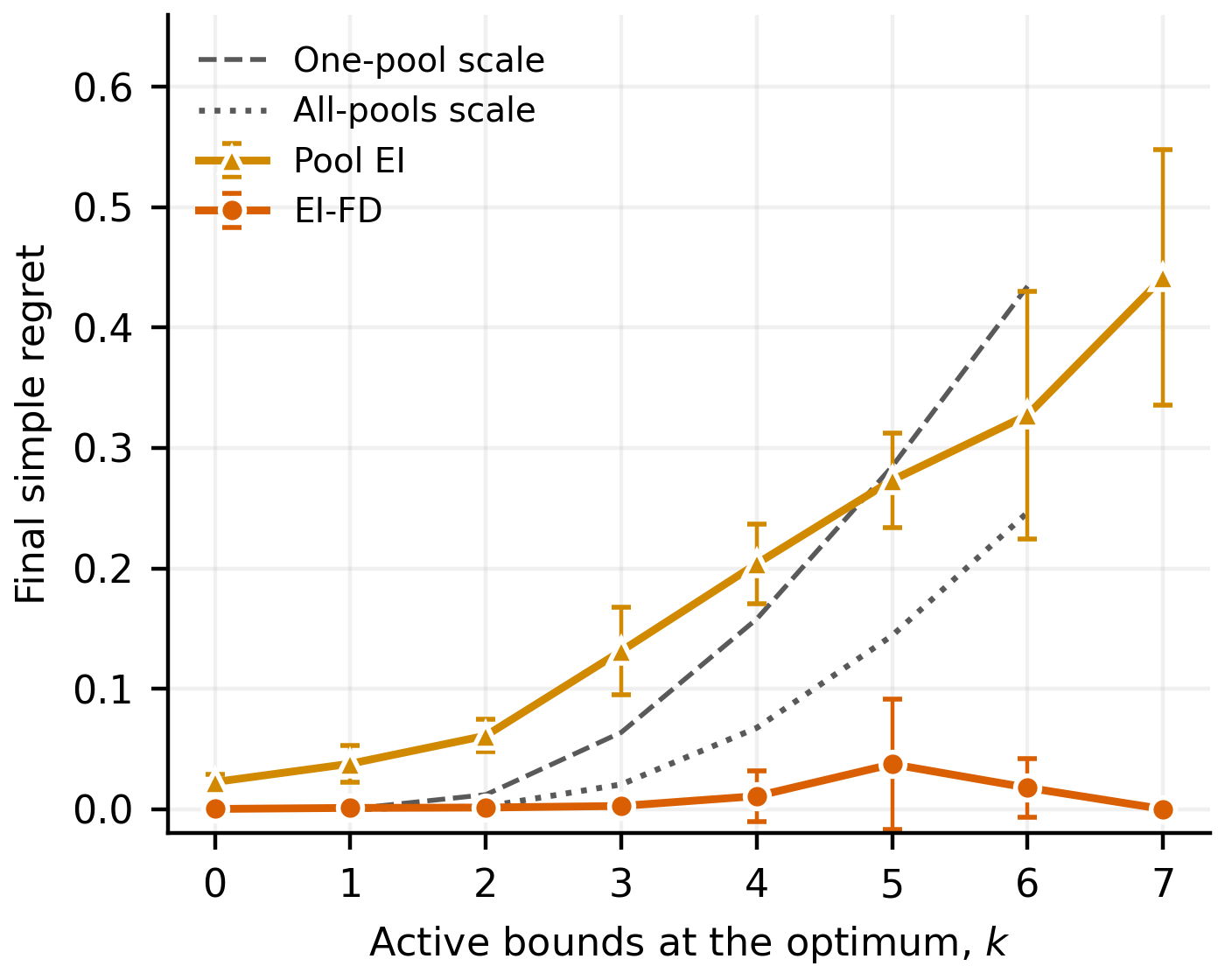}
    \caption{Final simple regret on the synthetic function of
    Eq.~\eqref{eq:synthetic_function} after 25 initial and 30 sequential
    evaluations, mean \(\pm\) sample SD over eight seeds. The parameter $k$ controls which quadratic centers lie outside the box. Dashed and dotted curves are
    the characteristic pool scales of Eq.~\eqref{eq:pool_floor} with \(\lambda=0.6\).}
    \label{fig:active_constraint_scaling}
\end{figure}

Figure~\ref{fig:active_constraint_scaling} reports the synthetic study of
Section~\ref{sec:scaling_protocol}. With the surrogate, budget, and pool size
held fixed, the final regret of pool search increases monotonically with the
number of active bounds, from \(0.023\) at \(k=0\) to \(0.131\) at \(k=3\),
\(0.273\) at \(k=5\), and \(0.442\) at \(k=7\). For \(k=5,6\) the measured
regret lies between the two reference scales from Eq.~\eqref{eq:pool_floor}: the
single-pool scale \(\lambda(k!/N)^{1/k}\) and the scale
\(\lambda(k!/TN)^{1/k}\) that assumes all \(T=30\) pools contribute. At
\(k=5\) these are \(0.285\) and \(0.144\). For small \(k\) the regret is above
both scales by about \(0.02\)--\(0.07\). This excess is consistent with
unresolved interior coordinates and adaptive search effects, which the
active-coordinate calculation does not separate. No pool run places any coordinate exactly on an active bound.

The pool-size dependence at \(k=5\) is consistent with the weak power-law scale. The
final regret is \(0.396\), \(0.273\), and \(0.156\) for \(N=500\), \(5000\),
and \(50\,000\), that is, factors of \(0.69\) and \(0.57\) per tenfold
increase against the predicted \(10^{-1/5}=0.63\). A hundredfold larger pool
therefore leaves \(39\%\) of the shortfall.

Continuous refinement substantially reduces the observed dependence on \(k\). Its final regret stays
below \(0.04\) for every \(k\) and is at most \(0.003\) for \(k\leq3\) and
zero to numerical precision at \(k=7\); the largest value,
\(0.037\pm0.054\) at \(k=5\), occurs in runs that have not yet
identified all active bounds after 30 acquisitions and is seven times smaller
than the corresponding pool regret. The coil designs also approach several box bounds and analytic constraints
(Appendix~\ref{sec:fixed_budget_results}). This is qualitative agreement
with the coverage diagnosis; counting nonlinear active constraints does
not substitute for $k$ in the uniform-box calculation.

\FloatBarrier
\section{Matched acquisition-search experiments}
\label{sec:optimization_application}
\subsection{Problems, initialization, and budgets}
The main comparison uses two deterministic FEM objectives with inexpensive
analytic feasibility checks. The seven-variable coil problem maximizes
surface-averaged magnetic-flux-density magnitude over a prescribed disk
below an axisymmetric coil--core assembly. Its constraints are box bounds,
core mass at most 200 g, copper mass at most 50 g, axial fit, and an aperture
inside the lower core radius. Production excitation fixes current density,
so total current changes with coil cross-section. The complete formulation,
material model, and mesh settings are in Appendix~\ref{sec:optimization_formulation}.
The independent six-variable heat-spreader problem minimizes flux-weighted
thermal resistance on a $30\times20$-mm plate. Three copper-rib widths and
heights vary within box bounds and a 100-mm$^2$ incremental copper-area budget.
Its conforming linear-triangle solver and boundary conditions are specified
in Appendix~\ref{sec:v3_heat_protocol}.

Each main experiment uses ten paired seeds, 42--51. All five policies share
25 initial observations within a seed, followed by 50 policy-specific
evaluations. Coil initialization uses a fixed 50-record holdout (split seed
42) from the existing 250-record feasible, LHS-generated dataset. Each run
then takes a seed-specific random subset of 25 records from the remaining
200, independently of their response values. These subsets need not retain
Latin-hypercube stratification and can overlap across seeds. Thermal
initialization uses 25 feasible points obtained by filtering and truncating
LHS-generated batches, evaluated once per seed and shared across policies.
All post-initialization coil observations are new FEM solves; unused table
entries are never queried by the online policies.

\subsection{Surrogate and acquisition}
Each policy fits its own zero-mean Mat\'ern-5/2 ARD GP after each observation,
with training-only input and target standardization, fitted signal and
white-noise variances, and eight hyperparameter optimizer restarts.
Acquisition uses the latent-function posterior variance, retaining the
fitted nugget in the training covariance. Hyperparameters and transformations
are detailed in Appendix~\ref{app:gp}; Section~\ref{sec:v3_baselines} describes
the transfer to differentiable tensor algebra. Every reported incumbent is an observed feasible FEM
value. Held-out predictive accuracy is reported separately and is not used
to select online policies.

For incumbent \(y_{\mathrm{best}}^{(n)}=\max_{i\leq n}y_i\), the standard
maximization-form expected improvement is
\begin{equation}
\begin{aligned}
\operatorname{EI}_n(\mathbf{u})
&=
\Delta_n(\mathbf{u})\Phi(z_n)
+\sigma_n(\mathbf{u})\phi(z_n), \\
\Delta_n
&=
\mu_n(\mathbf{u})-y_{\mathrm{best}}^{(n)}, \\
z_n
&=
\frac{\Delta_n}{\sigma_n(\mathbf{u})}.
\end{aligned}
\label{eq:compact_expected_improvement}
\end{equation}
with EI set to zero when \(\sigma_n(\mathbf{u})=0\). The exploration offset is
\(\xi=0\). Acquisition uses the latent-function predictive standard
deviation, excluding the fitted white-noise contribution.

In the pool-based reference, denoted pool EI, EI is optimized by
deterministic finite-candidate maximization rather than a continuous or
multi-start optimizer. At online iteration \(n\), a new pool of
5000 feasible candidates from LHS-generated batches is generated in \([0,1]^7\) using seed
\(s+1009n\). Analytic geometric and mass constraints are applied during pool
generation. Each batch contains $\max(2N,64)$ LHS points; feasible
points are retained until $N=5000$ are available, then truncated.
Filtering and truncation do not preserve exact LHS stratification. Candidates whose Euclidean distance from any observed normalized
design is at most \(10^{-3}\) are removed. The GP mean, standard deviation,
and EI are then evaluated for every remaining candidate, and the point with
the largest EI is selected; ties follow NumPy's first-maximum rule. No
continuous acquisition optimizer or restart procedure is used in this
reference. The selected
design is mapped to physical units, evaluated once by COMSOL, appended to the
run-specific dataset, and used to refit the GP. This continues for 50
sequential acquisitions, giving 75 accounted FEM observations per run. Final
designs and best-so-far trajectories use only observed FEM responses, never
unevaluated GP predictions.

\subsection{Continuous Constrained Refinement}
\label{sec:constraint_aware_search}

The finite-difference refinement keeps the GP, the acquisition function, and the candidate
pool of the reference method, and adds a constrained refinement. At iteration
\(n\), with acquisition \(\alpha_n\) and the pool of the previous subsection:
\begin{enumerate}
\item evaluate \(\alpha_n\) on the pool and retain the five best candidates
together with the current incumbent design as starting points;
\item from each start, maximize \(\alpha_n(\mathbf{u})\) with sequential
least-squares programming subject to \(\mathbf{0}\leq\mathbf{u}\leq\mathbf{1}\)
and the analytic inequalities of
Eq.~\eqref{eq:compact_optimization_objective} (axial fit, core mass, copper
mass, and \(r_1\geq c_1\)), with EI scaled by the pool maximum and objective tolerance $10^{-10}$,
using forward-difference gradients with step
\(10^{-5}\) that are reversed at an upper bound, and at most 60 iterations;
\item verify each result with the same feasibility function that generates the
pool; a result that violates it is pulled back along the segment to its
feasible start by bisection;
\item discard results closer than \(10^{-3}\) to an observed design, and
simulate the candidate with the largest acquisition value among the pool
maximum and the refined points.
\end{enumerate}
The selected acquisition value is never lower than the pool maximum, every
simulated design is feasible, and the refinement requires no FEM call. Its
cost was approximately \(0.2\)--\(0.25\) s per iteration in the present implementation.

All continuous policies enforce the same analytic constraints. FD denotes
finite-difference derivatives and AD automatic derivatives. Two acquisition
functions are run with the finite-difference search: EI of
Eq.~\eqref{eq:compact_expected_improvement} (EI-FD) and the
GP posterior mean \(\mu_n\) (Mean-FD). The second omits the
uncertainty term and serves as a prospective online control that separates
the effect of the acquisition search from that of uncertainty-aware
acquisition. Both use the same paired seeds, initial designs, pool seeds, and GP
settings as pool EI: ten seeds in the main comparison and twenty in
the extended coil study.

\subsection{Matched Continuous-Acquisition Baselines}
\label{sec:v3_baselines}
The automatic-gradient controls separate the geometric limitation of a
finite candidate pool from the numerical implementation of continuous
acquisition maximization. They use the ten paired seeds, 42--51, and the shared initialization, evaluation
budget, candidate-pool generation, and GP specification described above. No
hyperparameters are fitted a second time during posterior transfer. Each
policy refits the GP to its own accumulated observations at every step; the fitted
training inputs, Cholesky factor, kernel parameters, and conditioning weights
are transferred to double-precision differentiable tensor algebra. Prediction
and derivative checks precede the expensive experiments.

The first control maximizes ordinary EI with exact automatic derivatives of
this posterior. The second maximizes numerically stable logarithmic EI using
the implementation of the log-EI helper in BoTorch
\citep{balandat2020botorch,ament2023logei}. BoTorch supplies the stable acquisition calculation;
SciPy SLSQP supplies the constrained optimizer in both controls. The transferred posterior retains the fitted scikit-learn GP parameters. EI is scaled by its largest candidate-pool value before optimization
to prevent physical output units from determining the stopping tolerance.
The logarithmic criterion does not require this rescaling.

Each control receives the same five highest-scoring pool starts and the
current incumbent, with at most 60 SLSQP iterations per start and an objective
tolerance of $10^{-10}$. They also receive the same box and analytic
constraints, feasible-segment pullback, $10^{-3}$ duplicate exclusion, and
pool-winner fallback as the finite-difference refinement. These shared safeguards prevent
a discrepancy between the feasibility tolerance of a nonlinear optimizer and
the exact FEM admission check from handicapping a baseline. The controls
therefore test whether changing derivatives or the numerical representation
of EI changes the attainable design after continuous, constraint-aware search
has been introduced. Every trial actually sent to FEM is recorded; failed
inner optimizations consume acquisition time but no FEM observation.

\subsection{Comparison and numerical verification}
We report terminal mean and sample standard deviation over the ten paired
seeds and pointwise 95\% Student-$t$ intervals on best-so-far trajectories.
The LogEI versus EI-FD comparisons use unadjusted two-sided Wilcoxon
signed-rank tests. Differences with magnitude at most $10^{-12}$ in the
stored objective units (T for coil, K/W for thermal) are set to zero for
win counts and these tests, and zero differences are omitted from ranking.
We adopted this tie rule during revision after finding that a roughly
$2\times10^{-19}$ T difference had been counted as a coil win. Using
unrounded stored observations without the rule gives nine LogEI wins, no
ties and one loss ($p=0.0273$); applying the rule gives eight wins, one tie
and one loss ($p=0.0391$). The archive verification script reads the original
per-seed observation files with round-trip precision; the summary CSV alone
does not preserve this difference. These tests are distinct from the Holm-adjusted comparisons
in the twenty-seed COMSOL study (Appendix~\ref{app:extended}).
All five main policies spend their continuation budgets on feasible designs.
Inner optimization failures cost acquisition time but no FEM evaluation.

The strongest observed coil designs are freshly re-solved at finer meshes,
larger outer domains, and their joint setting. Each thermal run's terminal
winner is re-evaluated at two finer meshes. These checks compare the same
designs across numerical settings. They establish sensitivity of the selected
designs, rather than trajectories from re-optimization on a finer mesh.
The full protocols and coil values appear in
Appendix~\ref{sec:v3_validation_protocol}.

Two development groups were excluded from all final summaries: incomplete
pilot runs with asymmetric feasibility admission, and thermal runs preceding
the coincident-interface mesh repair. Final matched runs use the common
admission safeguards and repaired mesh implementation. The supplement records
these reasons in its protocol and contains the final reported histories.

\section{Results on electromagnetic and thermal design}
\label{sec:results_discussion}
\subsection{Prospective Continuous-Acquisition Controls}
\label{sec:v3_baseline_results}
Table~\ref{tab:v3_controls} and Fig.~\ref{fig:v3_coil} compare ten matched seeds (42--51). The twenty-seed coil comparison
is reported in Appendix~\ref{sec:fixed_budget_results}. The controls use the same GP specification, initial observations, start
rule, constraints, and admission safeguards. Each policy subsequently
refits its own GP to its own observations.
EI-FD reaches 0.9745 $\pm$ 0.0019 mT, compared with 0.9120 $\pm$ 0.0136 mT for pool EI. Continuous EI with automatic derivatives reaches 0.9721 $\pm$ 0.0035 mT and continuous LogEI reaches 0.9770 $\pm$ 0.0023 mT.
LogEI has the highest mean coil response among the EI-based policies and
beats finite-difference EI-FD on eight seeds, ties on one, and loses on one
(unadjusted two-sided Wilcoxon $p=0.0391$). GP mean has the highest nominal coil mean
across all five policies, at 0.9780 mT. The small differences among the
continuous policies are comparable to the observed numerical sensitivity;
they do not establish a resolved ranking of physical designs. The dominant response gain comes from replacing the finite pool with
continuous feasible search; LogEI provides the strongest tested EI-based
implementation on the production mesh.

\begin{table}[!tb]
\centering\small
\caption{Matched prospective acquisition controls at 75 FEM observations,
mean $\pm$ sample SD over ten seeds. Higher coil response and lower thermal
resistance are better. The coil values for the first three policies are
recomputed from the same ten archived seed histories.}
\label{tab:v3_controls}
\begin{tabular}{lcc}\toprule
Policy & Coil ROI response (mT) & Thermal resistance (K/W)\\\midrule
Pool EI & 0.9120 $\pm$ 0.0136 & 12.688 $\pm$ 0.173 \\
EI-FD & 0.9745 $\pm$ 0.0019 & 11.905 $\pm$ 0.026 \\
Mean-FD & 0.9780 $\pm$ 0.0003 & 11.903 $\pm$ 0.152 \\
EI-AD & 0.9721 $\pm$ 0.0035 & 11.900 $\pm$ 0.045 \\
LogEI-AD & 0.9770 $\pm$ 0.0023 & 11.860 $\pm$ 0.011 \\
\bottomrule\end{tabular}
\end{table}
\begin{figure}[!tb]\centering
\includegraphics[width=.65\textwidth]{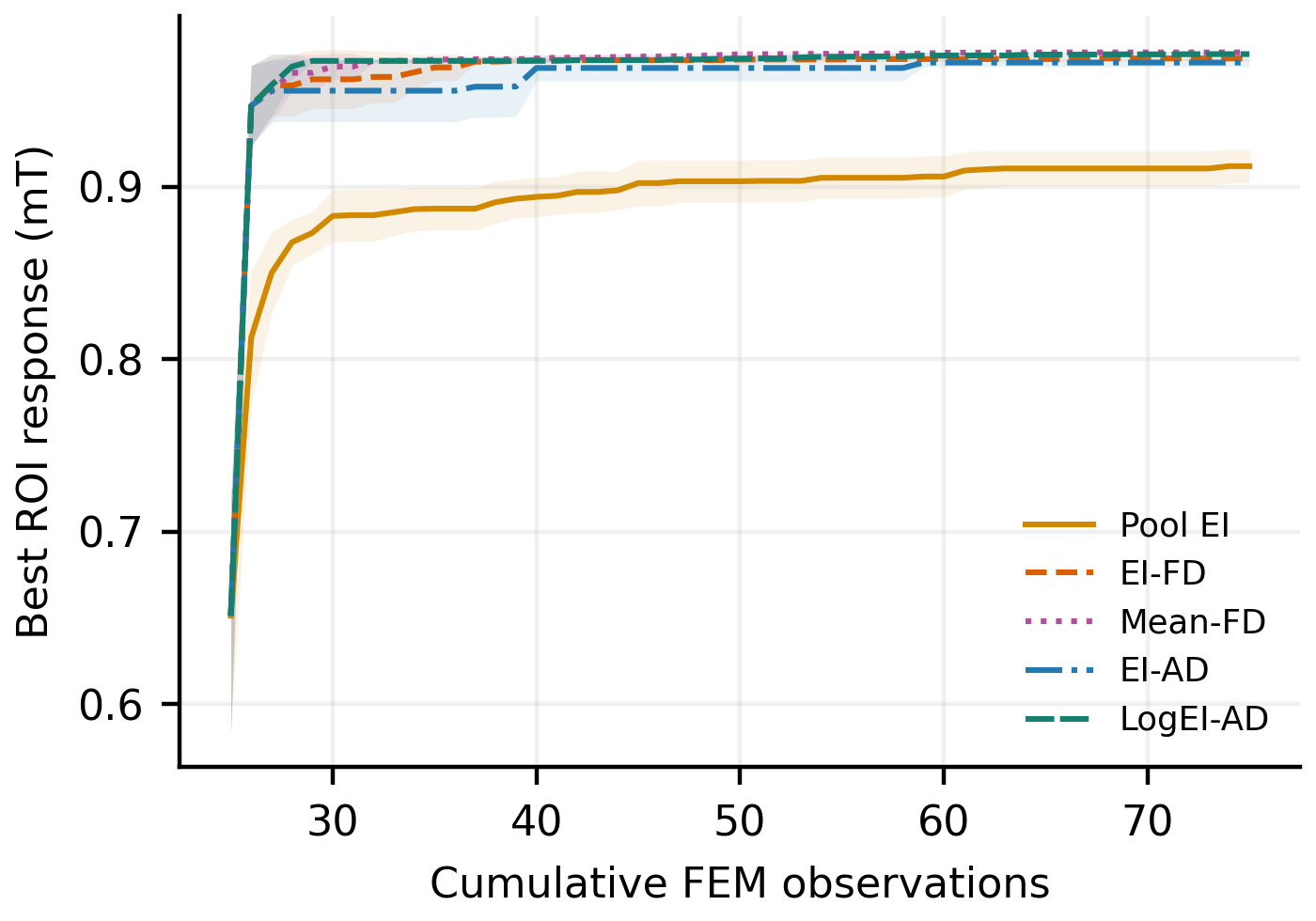}
\caption{Matched ten-seed coil comparison including continuous EI and LogEI.
Bands are pointwise 95\% Student-$t$ intervals.}
\label{fig:v3_coil}\end{figure}

\subsection{Transfer to Independent Conduction FEM Design}
\label{sec:v3_heat_results}
On the independent conduction problem, EI-FD reduces mean terminal resistance from 12.688 $\pm$ 0.173 K/W for pool EI to 11.905 $\pm$ 0.026 K/W, a 6.2\% change in the means. Mean-FD reaches 11.903 $\pm$ 0.152 K/W. EI-AD reaches 11.900 $\pm$ 0.045 K/W. LogEI-AD reaches 11.860 $\pm$ 0.011 K/W. The paired EI-FD versus pool comparison favors the former on 10/10 seeds.
LogEI has the lowest mean resistance of all five policies and beats
EI-FD on 10/10 seeds (unadjusted two-sided Wilcoxon
$p=0.00195$). Its mean advantage is 0.0444 K/W, about 0.37\%.
The GP-mean average conceals an outlier on seed 51: 12.336 K/W, versus about
11.85 K/W on its other seeds. This larger spread prevents extending the
coil's exploitation-only conclusion to the thermal benchmark. It does not,
by itself, prove that the uncertainty term is the only cause of the failure.

\begin{figure}[!tb]\centering
\includegraphics[width=.85\textwidth]{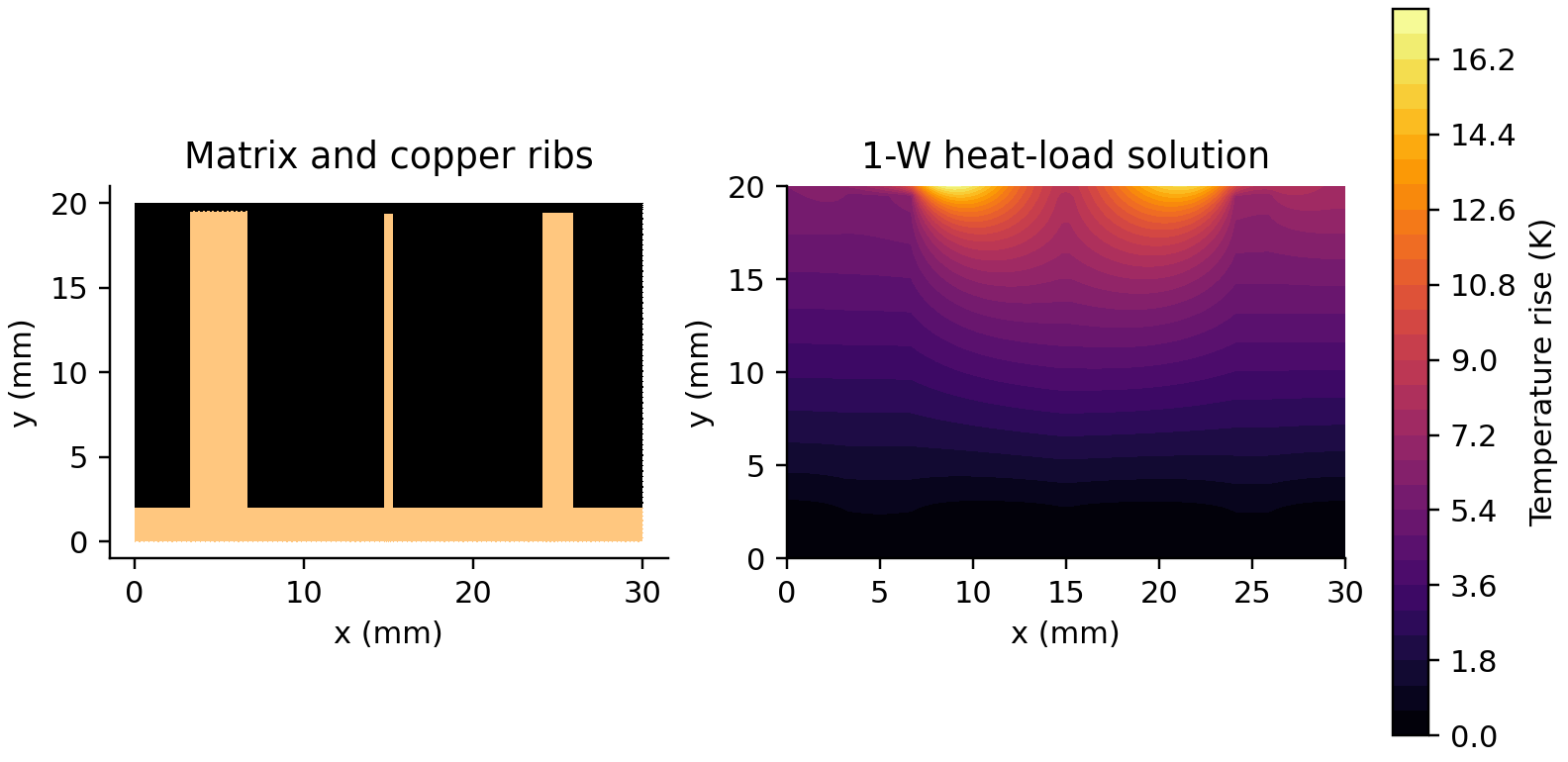}
\caption{Representative seed-42 EI-FD conduction design,
re-solved on the four-times-refined mesh. Left: matrix and copper layout.
Right: temperature rise with a 1-W applied load and zero-rise bottom.}
\label{fig:v3_heat_field}\end{figure}

All ten EI-FD terminal designs use at least 99.98\% of the incremental copper budget, with rib heights between 17.374 and 17.500 mm against the 17.5-mm upper bound. The independent designs therefore also concentrate near intersecting material and geometric limits. Re-solving the same terminal designs on the four-times-refined mesh gives 12.285 $\pm$ 0.032 K/W for EI-FD and 13.065 $\pm$ 0.174 K/W for pool EI. The refinement raises the EI-FD mean by about 3.2\%, while
the gap to pool EI remains about 6.0\%. Resistance is still changing at the
finest mesh, so these are sensitivity checks rather than a mesh-converged
reference. The finest-mesh LogEI mean is 12.242 K/W, so its nominal advantage
persists under the tested refinements. The 0.37\% production-mesh LogEI advantage is smaller than this
absolute mesh sensitivity; a mesh-converged physical advantage is not
established. Low algebraic residuals and heat-balance errors verify the
discrete solve, not its continuum accuracy.
Across all 150 mesh checks the largest heat-balance error is $3.5\times10^{-8}$ W and the largest free-node residual is $3.7\times10^{-9}$ W. The uniform-slab patch test reproduces $44.444444$ K/W, its analytic
value, to numerical precision. Figure~\ref{fig:v3_heat_field} shows the
material interfaces and temperature field of a representative EI-FD
design. This independently solved problem extends the coverage diagnosis to
conduction with six design variables and an active material-allocation rule.

\begin{figure}[!tb]\centering
\includegraphics[width=.65\textwidth]{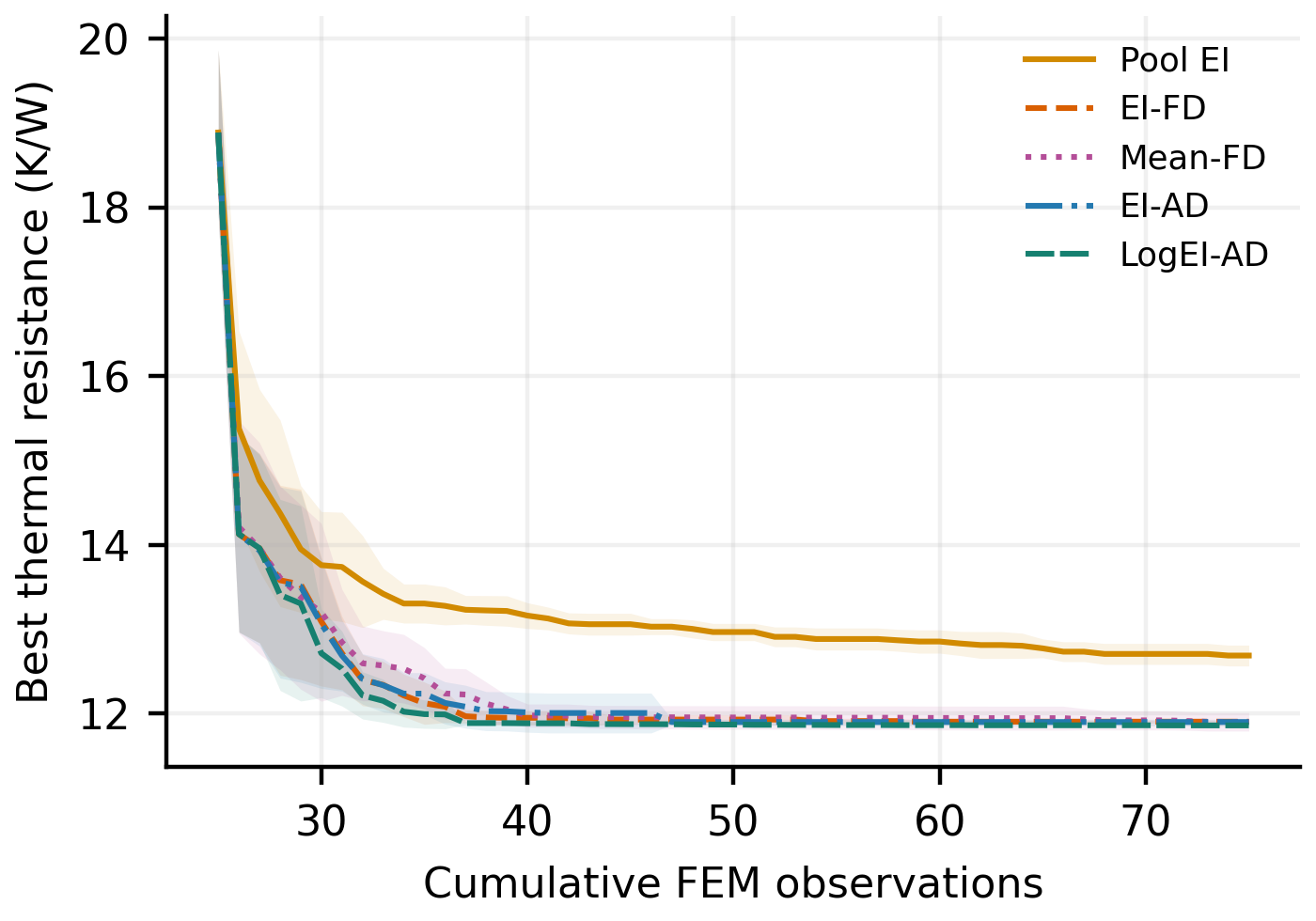}
\caption{Independent conduction FEM: best thermal resistance over ten paired
seeds. Lower is better; bands are pointwise 95\% Student-$t$ intervals.}
\label{fig:v3_heat}\end{figure}

\subsection{Boundary-enriched pool control on thermal design}
\label{sec:thermal_boundary_control}
An additional experiment tests whether boundary enrichment helps sequential
EI search on the thermal problem. We fixed the sampling rule before running
all ten seeds (42--51), reused their 25 initial observations, and retained the
same GP fitting, 50 additional solves and 5000 candidate slots per iteration.
Half the slots retain the first 2500 candidates of the original feasible pool.
For the other half, LHS proposals have a uniformly chosen number of coordinates
from one to six set to independently sampled zero or one endpoints. Infeasible
transformed points are rejected and replenished. Repeated vertices count as
candidate slots; the original distance-to-observations filter is then applied.
EI selects the best remaining candidate without local refinement. This rule
uses neither objective values nor knowledge of the optimum during generation.

Boundary enrichment reduces mean terminal resistance to $12.092\pm0.084$ K/W,
compared with $12.688\pm0.173$ for the original pool. The paired reduction is
$0.597$ K/W (95\% Student-$t$ interval $[0.484,0.709]$), positive on all ten
seeds. It remains worse than EI-FD by $0.187$ K/W and LogEI-AD by $0.232$ K/W,
also on every seed (each unadjusted two-sided Wilcoxon $p=0.00195$).
Re-evaluating the ten selected designs on the four-times-refined mesh gives
$12.465\pm0.078$ K/W, between the original pool's $13.065\pm0.174$ and
EI-FD's $12.285\pm0.032$ K/W. Thus boundary enrichment recovers part of the
pool-search deficit on this nonlinear feasible region, while continuous
refinement retains an advantage. These exploratory comparisons concern one
fixed enrichment rule and do not establish the best discrete candidate policy.

\subsection{Supporting coil checks}
The joint finest-mesh/largest-domain re-solves preserve the continuous-search
advantage over the selected pool winner. They can reverse the nominal
ordering of close continuous policies, which is why Table~\ref{tab:v3_controls}
is interpreted as production-mesh performance. The twenty-seed coil study
and fixed-25-A re-optimization also favor EI-FD over pool EI; the latter
changes the active set and reduces the terminal mean gap to about 1.5\%.
Appendices~\ref{app:extended} and \ref{sec:fixed_current_protocol} report these
additional comparisons, including the unequal use of initial observations
and infeasible evaluations by COMSOL-embedded optimizers.

\section{Scope and practical implications}
\label{sec:limitations_results}
The sampling calculation assumes independent uniform box candidates, positive
active multipliers, and a local linear loss. It does not prove a universal
regret bound for LHS, nonlinear constraints, adaptive BO, or interior optima.
The synthetic study tests a constructed family; two engineering models do
not establish performance over a broad benchmark distribution.

The coil uses one axisymmetric magnetostatic geometry, two excitation
formulations, and a fixed ROI. It is not a fabricated-device experiment,
nor a fixed-power or electrical-efficiency optimization. The conduction
problem has independent code and physics but prescribed material properties
and boundary conditions. Its production mesh is coarse, and no converged
continuum optimum is claimed. Re-solving terminal designs does not establish
the optimization trajectories that would occur on finer meshes.

The standard continuous controls use ten seeds, whereas the extended coil
comparison uses twenty. Embedded optimizers receive the shared incumbent and
an interior start, but do not fit all 25 initial observations as the GP
policies do. They also spend some evaluations on infeasible trial designs.
Their ranking is therefore specific to these implementations and accounting
rules. They were not run on the fixed-current or thermal problems; the
automatic-gradient controls were not run on fixed current either.

LogEI is the strongest tested EI-based policy in the matched production-mesh
comparisons, not a universally best optimizer. GP mean leads nominally on
the coil but has a thermal outlier. Numerical sensitivity prevents turning
the small continuous-policy differences into claims of resolved physical
superiority. Simulation-dependent feasibility, strongly multimodal responses,
three-dimensional models, and experimental validation remain open.

\FloatBarrier
For a new design problem, inspect whether high-scoring candidates and
observed incumbents approach several feasible boundaries. If so, test whether
continuous constrained acquisition search improves the attainable acquisition
value under the same model. Boundary-enriched candidate sets offer another
coverage check: they improve the multi-bound synthetic cases and recover part
of the thermal pool-search deficit, but lose efficiency on interior synthetic
optima. Broader performance on nonlinear feasible regions remains untested. Keep the acquisition
criterion, GP specification, and FEM admission rules matched when attributing
an observed improvement to the search procedure.

\section{Conclusion}
\label{sec:conclusion_future_work}
Finite random candidate pools can undersample good designs near intersecting
active bounds. The uniform-box calculation identifies an $N^{-1/k}$
active-gap scale, and the oracle experiment shows a coverage deficit even
when candidates are ranked by the true objective. Boundary enrichment reduces
that deficit in the multi-bound cases while exposing a tradeoff for interior
optima. Matched electromagnetic and thermal experiments then show that
continuous feasible acquisition search makes better use of the FEM budget
than the specified random pools. Existing continuous EI and stable LogEI
provide effective implementations. The practical lesson is to check whether
acquisition search can reach high-value feasible boundaries before changing
the surrogate or spending more simulations.

\section*{Data and code availability}
A separate reproducibility archive has been prepared for journal review. It
contains experiment code, recorded trajectories, numerical checks, protocols,
package versions, and scripts to recompute the reported summaries. This
archive, including the COMSOL model and numerical CSV files, is not included
in this arXiv version and has not been publicly released. The independent
thermal solver and synthetic experiments run without COMSOL; coil
re-evaluation requires licensed COMSOL 6.3, MPh, and the source model.
Appendix~\ref{app:numerical-records} describes the separately retained records.

\section*{Broader impact}
The experiments study numerical design optimization under prescribed material
and geometric constraints. More effective acquisition search can reduce
simulation use, but optimizing a simulated field or thermal response does
not establish device safety or operating limits. Physical deployment requires
application-specific validation, including electromagnetic exposure, heating,
material uncertainty, and manufacturing tolerances as appropriate. The coil
results are not evidence of clinical efficacy or suitability for human use.

\section*{LLM Usage Statement}
We used large language models (OpenAI's ChatGPT and
Anthropic's Claude) in a limited, assistive way while
preparing this work. The full manuscript, or major parts
of it, was given to these models to get reviewer-style feedback.
LLMs were also used to edit and refine text the authors
wrote (language, clarity and structure), to edit and refine
code the authors wrote, and to refine figures and visual
material the authors made. LLMs did not generate research
ideas, hypotheses, claims or conclusions. They also played
no part in designing, running or analyzing experiments, in
creating or annotating data, or in preparing related work and
citations. The authors reviewed and verified every change
an LLM suggested and take full responsibility for all submitted content.

\section*{Acknowledgments}
The authors thank Prof. Tao Song at the Institute of Electrical Engineering,
Chinese Academy of Sciences, for guidance and discussions during a previous
summer research project on electromagnetic simulation.

\bibliography{references}
\bibliographystyle{tmlr}
\clearpage
\appendix
\section{Electromagnetic benchmark specification}
\subsection{Optimization Problem Formulation}
\label{sec:optimization_formulation}

The benchmark maximizes the FEM-observed surface-averaged magnetic-flux-density
magnitude over a prescribed region of interest (ROI). With
\(\mathbf{x}=[c_1,r_1,t_1,l_1,l_2,w_1,h_1]^{\mathsf T}\), the complete problem
is
\begin{equation}
\begin{aligned}
\max_{\mathbf{x}}\quad & B_{\mathrm{ROI}}(\mathbf{x}) \\
\text{s.t.}\quad
& M_{\mathrm{core}}(\mathbf{x})\leq200~\mathrm{g}, \\
& M_{\mathrm{coil}}(\mathbf{x})\leq50~\mathrm{g}, \\
& l_1-h_1-2~\mathrm{mm}\geq0, \\
& r_1-c_1\geq0, \\
& \mathbf{x}^{\min}\leq\mathbf{x}\leq\mathbf{x}^{\max}.
\end{aligned}
\label{eq:compact_optimization_objective}
\end{equation}
The bounds are listed in Table~\ref{tab:design_vars}. The two mass limits are
kept separate because magnetic-core material and copper winding are not treated
as interchangeable resources. The 50-g copper limit is a protective cap: within the present box
bounds its analytic maximum is approximately 31.6~g, so it is not active in the
reported runs. A single 250-g total-mass limit would define a different feasible
domain by allowing core and copper allocations to compensate for one another;
that alternative problem is not evaluated here. These limits are benchmark
design choices rather than universal manufacturing limits.

The excitation is held at fixed current density rather than fixed total current.
Consequently, the objective measures the response of the combined geometry and
its area-dependent applied current; it is not an electrical-efficiency
objective. Appendix~\ref{sec:fem_setup} defines the ROI response and physical
model in detail.

\subsection{Parameterized Coil--Core Benchmark}

\begin{table}[!tb]
\centering
\caption{Independent design variables defining the parameterized benchmark geometry.}
\label{tab:design_vars}
\small
\begin{tabular}{llll}
\toprule
Symbol & Description & Unit & Feasible range \\
\midrule
$c_1$ & Aperture radius & mm & $[0.5,\,5]$ \\
$r_1$ & Lower core radius & mm & $[10,\,20]$ \\
$t_1$ & Radial extension of upper core & mm & $[0.5,\,10]$ \\
$l_1$ & Lower core height & mm & $[2.5,\,20]$, with $l_1\geq h_1+2$ \\
$l_2$ & Upper core height & mm & $[0.5,\,20]$ \\
$w_1$ & Coil radial width & mm & $[0.5,\,5]$ \\
$h_1$ & Coil axial height & mm & $[0.5,\,5]$ \\
\bottomrule
\end{tabular}
\end{table}

\begin{table}[!tb]
\centering
\caption{Fixed physical and material parameters for the benchmark.}
\label{tab:consts}
\small
\begin{tabular}{llll}
\toprule
Symbol & Description & Unit & Value / Assumption \\
\midrule
$\rho_{\mathrm{Cu}}$ & Copper density & kg\,m$^{-3}$ & 8940 \\
$\sigma_{\mathrm{Cu}}$ & Coil electrical conductivity & S\,m$^{-1}$ & \(6.0\times10^7\) \\
$J_{\mathrm{apply}}$ & Applied current-density parameter & A\,m$^{-2}$ & \(1.0\times10^6\) \\
$\rho_{\mathrm{core}}$ & Core density & kg\,m$^{-3}$ & 7000 \\
$B_{\mathrm{sat}}$ & High Flux saturation flux density & T & 1.5 \citep{magneticsHighFlux} \\
Grade & Manufacturer nominal permeability grade & -- & 125 \citep{magneticsHighFlux} \\
$\mu_{r,\mathrm{sec}}$ & Low-field secant value implied by the first table interval & -- & 120.0 \\
$\mu_{r,\mathrm{air}}$ & Air relative permeability & -- & 1 \\
\bottomrule
\end{tabular}
\end{table}

\begin{table}[!tbp]
\centering
\caption{Derived geometric and screening quantities.}
\label{tab:derived}
\small
\begin{tabular}{llll}
\toprule
Symbol & Description & Unit & Expression \\
\midrule
$r_2$ & Upper core radius & mm & $r_1+t_1$ \\
$r_0$ & Coil outer radius & mm & $r_1+w_1$ \\
$M_{\mathrm{core}}$ & Core mass & kg &
$\rho_{\mathrm{core}}V_{\mathrm{core}}$ \\
$M_{\mathrm{coil}}$ & Coil mass & kg &
$\rho_{\mathrm{Cu}}V_{\mathrm{coil}}$ \\
$M_{\mathrm{total}}$ & Total mass & kg &
$M_{\mathrm{core}}+M_{\mathrm{coil}}$ \\
$I_{\mathrm{coil}}$ & Applied coil current & A &
$J_{\mathrm{apply}}w_1h_1$ \\
\bottomrule
\end{tabular}
\end{table}

Tables~\ref{tab:design_vars}--\ref{tab:derived} summarize the independent
design variables, fixed physical parameters, and analytically derived
screening quantities, respectively.

The controlled benchmark considered in this work is a parameterized
current-excited coil--core geometry for localized magnetic-field evaluation.
Its computational geometry consists of a stepped soft-magnetic core domain and a
concentric copper coil domain, with a prescribed region of interest
(ROI) beneath the assembly. An axial aperture is included while preserving
geometric symmetry.
The model is not presented as a validated representation of a particular
commercial material or device. Instead, it provides a fixed and sufficiently
structured reference problem for evaluating surrogate-assisted search.

\begin{figure}[!tb]
\centering
\begin{tikzpicture}[x=0.78cm,y=0.78cm,font=\scriptsize]
\draw[->] (0,-1.05) -- (0,3.05) node[above] {\(z\)};
\draw[->] (0,-0.9) -- (4.3,-0.9) node[right] {\(r\)};
\filldraw[fill=gray!35,draw=black] (0.55,0) rectangle (2.45,1.15);
\filldraw[fill=gray!35,draw=black] (0.55,1.15) rectangle (3.15,2.35);
\draw[thick,dashed,fill=blue!8] (2.45,0.2) rectangle (3.55,1.0);
\draw[very thick,red] (0,-0.55) -- (1.05,-0.55);
\node[red,below] at (0.55,-0.55) {ROI};
\node at (1.45,1.75) {High Flux core};
\node[align=center] at (3.05,0.55) {copper\\coil};
\draw[<->] (0,-0.05) -- (0.55,-0.05) node[midway,below] {\(c_1\)};
\draw[<->] (0,2.55) -- (2.45,2.55) node[midway,above] {\(r_1\)};
\draw[<->] (2.45,2.55) -- (3.15,2.55) node[midway,above] {\(t_1\)};
\draw[<->] (2.22,0) -- (2.22,1.15) node[midway,left] {\(l_1\)};
\draw[<->] (3.35,1.15) -- (3.35,2.35) node[midway,right] {\(l_2\)};
\draw[<->] (2.45,0.08) -- (3.55,0.08) node[midway,below] {\(w_1\)};
\draw[<->] (3.75,0.2) -- (3.75,1.0) node[midway,right] {\(h_1\)};
\end{tikzpicture}
\caption{Meridional schematic of the axisymmetric benchmark
parameterization (not to scale). The dashed copper-coil domain carries the
current excitation and is included in the geometric and mass constraints.}
\label{fig:benchmark_geometry}
\end{figure}

The upper-core and coil outer radii are $r_2=r_1+t_1$ and
$r_0=r_1+w_1$. The coil starts at $z=2$ mm, giving the axial-fit constraint
$l_1\ge h_1+2$ mm. The core and coil masses are computed from their revolved
volumes. All screening and optimization stages use the same bounds and
design-space fingerprint. The coil is a homogenized annular current domain,
not an explicit turn-resolved winding. Inputs are normalized by
$u_j=(x_j-x_j^{\min})/(x_j^{\max}-x_j^{\min})$.

\subsection{Electromagnetic FEM Model}
\label{sec:fem_setup}
The two-dimensional axisymmetric model solves steady magnetostatics,
\begin{equation}
\mathbf B=\nabla\times\mathbf A,\qquad
\nabla\times\mathbf H(\mathbf B)=\mathbf J_e.
\end{equation}
The core is an isotropic High Flux 125 domain represented by an embedded
63-point DC $B$--$H$ table. The manufacturer's curves provide material-family
context \citep{magneticsHighFlux,magneticsHighFluxCurves}; row-level
digitization provenance is not independently preserved in the archive.
The first nonzero point, $(795.77\,\mathrm{A/m},0.12\,\mathrm{T})$, implies a
low-field secant relative permeability of about 120. The nominal grade 125
and quoted saturation level of 1.5 T are not remanent-field inputs.
The configured remanence is zero. Selected baseline designs have core maxima
below 0.064 T, within the first interpolation interval; those results do not
test nonlinear saturation. Hysteresis, eddy-current loss, and coupled thermal
effects are outside this stationary benchmark.

The meridional air domain spans $0\le r\le200$ mm and
$-100\le z\le100$ mm, with magnetic insulation
$\mathbf n\times\mathbf A=0$ on the outer boundary and axisymmetry on the
axis. COMSOL 6.3 rebuilds an unstructured triangular mesh using automatic
size level 3 after each geometry change. Mesh levels 2 and 1 and outer
domains of 300 and 400 mm are used for verification, including the joint
finest-mesh/largest-domain setting.

The ROI is the disk obtained by revolving $0\le r\le5$ mm at $z=-10$ mm.
Its scalar response is
\begin{equation}
B_{\rm ROI}=\frac{\int_{\Gamma_{\rm ROI}}|\mathbf B(r,z)|2\pi r\,ds}
{\int_{\Gamma_{\rm ROI}}2\pi r\,ds}.
\end{equation}
COMSOL's boundary-average probe applies the axisymmetric weighting; it is
not repeated in post-processing. The prescribed ROI is a benchmark choice,
not an application-derived performance standard.

The production excitation is $I_{\rm coil}=J_{\rm apply}w_1h_1$, with
$J_{\rm apply}=10^6$ A/m$^2$. Thus current grows with coil cross-section,
and the objective is field response, not electrical efficiency. Fresh source
reloads give zero ROI field at zero applied current. A separate full
re-optimization at fixed 25 A is described in
Appendix~\ref{sec:fixed_current_protocol}. Python and MPh automate geometry,
meshing, solving, and extraction without saving changes to the source model.

\FloatBarrier
\section{Independent thermal benchmark specification}
\subsection{Independent Heat-Spreader Design Benchmark}
\label{sec:v3_heat_protocol}
An independently implemented conduction problem tests transfer beyond
magnetostatics, COMSOL, and the coil parameterization. It is a numerical heat-spreader benchmark with prescribed materials and
boundary conditions; no fabricated heat sink is evaluated.
Consider a $30\times20$-mm rectangular plate of depth $d=1$ mm containing a
matrix of conductivity $15$ W/(m K), a fixed 2-mm-high copper base, and three
copper ribs of conductivity $400$ W/(m K). The ribs are centered at
$x=5,15,25$ mm and extend upward from the base. Their widths and heights are
the six design variables,
\begin{equation}
 \begin{gathered}
 \mathbf{x}_{\rm th}=(w_1,w_2,w_3,h_1,h_2,h_3),\\
 0.5\le w_j\le5,\quad 4\le h_j\le17.5,
 \end{gathered}
\end{equation}
where dimensions in the bounds are millimeters. An incremental copper-area
budget imposes $\sum_{j=1}^3 w_jh_j\le100$ mm$^2$; the fixed base is excluded
from this allocation. The ribs remain disjoint and inside the plate for all
admissible dimensions.

The temperature rise $\theta$ obeys steady conduction,
\begin{equation}
 -\nabla\cdot(\kappa(\mathbf{x}_{\rm th};\mathbf{r})\nabla\theta)=0.
\end{equation}
The bottom is held at zero temperature rise and the sides are insulated. An
inward heat flux $q(x)$ on the top is proportional to
\begin{equation}
 e^{-\frac12[(x-7.5)/2.5]^2}
 +0.65e^{-\frac12[(x-22.5)/3]^2},
\end{equation}
with $x$ in millimeters, normalized so that
$Q=\int_{\Gamma_{\rm top}}q\,d\,ds=1$ W.
Temperature and normal heat flux are continuous at material interfaces.
The performance metric is the flux-weighted thermal resistance
\begin{equation}
 R_{\rm th}=\frac{1}{Q^2}
   \int_{\Gamma_{\rm top}}q\theta\,d\,ds\quad[\mathrm{K/W}],
 \qquad \max f_{\rm th}=-R_{\rm th}.
\end{equation}
Thus a lower reported resistance is better. With $Q=1$ W the metric equals
the heat-load-weighted top temperature rise numerically.

The weak form is discretized with conforming three-node linear triangles.
All rib sides, rib tips, and the base interface are explicit mesh lines;
conductivity is assigned elementwise without a pixel approximation of the
moving interfaces. Numerically coincident interface coordinates within
$10^{-7}$ mm are coalesced before meshing to avoid ill-conditioned sliver
elements; an equal-tip versus nearly equal-tip regression check verifies
that this tolerance does not create a spurious design benefit.
Each interval between interface lines is subdivided to a
maximum coordinate spacing of 1.25 mm, and each rectangular cell is split
into two triangles. Three-point Gauss quadrature integrates top-boundary
loads. SciPy assembles and solves the sparse Dirichlet system; no COMSOL
model or coil data are used. A uniform-conductivity, uniform-load patch test
checks the exact resistance $H/(\kappa Wd)$, and the computed bottom reaction
checks energy conservation.

Ten seeds, 42--51, each generate 25 feasible initial designs from filtered LHS batches,
which are solved once and shared by all five policies. Pool EI,
EI-FD, Mean-FD, continuous automatic-gradient
EI, and continuous LogEI each receive 50 further FEM observations. The
5000-candidate pool rule, six-start refinement, duplicate threshold, GP
standardization, kernel bounds, and eight hyperparameter restarts follow the
coil protocol; only the dimension and physical evaluator change. Retained
histories include actual geometries, responses, GP-fit time, acquisition time,
FEM time, and inner-solver diagnostics. This small FEM problem evaluates
transfer and response per solve, not a wall-clock speedup on an expensive
industrial model.

\FloatBarrier
\section{Surrogate specification and held-out validation}
\label{app:gp}
\subsection{Gaussian-Process Surrogate}
The archived 250-record coil table retains physical and normalized inputs,
FEM responses, and feasibility metadata. Predictive validation is separate
from the prospective optimization histories.

\noindent\textbf{Preprocessing and Data Splits}

The 250 FEM records are represented by normalized coordinates
\(\mathbf{u}_i\in[0,1]^7\) and scalar responses \(y_i\). For each of five
random seeds, 50 records are held out before fitting and the remaining 200
form the training pool. Nested subsets of \(25,50,\ldots,200\) training
records are used for the learning curves, while the 50-sample validation set
remains fixed within each split.

Within every fit, each input coordinate and the target are standardized using
means and population standard deviations (\texttt{StandardScaler}, divisor
$n$) computed only from the active training subset. The same training-derived transformations are applied to validation
points and sequential-search candidates. Predictions are inverse-transformed
to tesla before computing metrics or expected improvement; validation
observations never enter preprocessing or hyperparameter estimation.

\noindent\textbf{Gaussian-Process Model}

The standardized response is modeled with a zero-mean Gaussian process
\citep{rasmussen2006gp} and a Mat\'ern-\(5/2\) covariance with automatic relevance determination
(ARD), plus a white-noise term:
\begin{equation}
\begin{aligned}
k(\mathbf{u},\mathbf{u}')
&=
\sigma_f^2
\left(1+\sqrt{5}d+\frac{5}{3}d^2\right)
e^{-\sqrt{5}d} \\
&\quad
+\sigma_n^2\delta_{\mathbf{u},\mathbf{u}'}, \\
d^2
&=
\sum_{j=1}^{7}
\frac{(\widetilde{u}_j-\widetilde{u}_j')^2}{\ell_j^2}.
\end{aligned}
\label{eq:compact_matern_ard}
\end{equation}
Here \(\widetilde{u}_j\) denotes an input standardized using the active
training subset. The recorded GP implementation uses Python~3.14.6, NumPy~2.4.3,
SciPy~1.17.1, and scikit-learn~1.8.0. The continuous-control environment
also records PyTorch~2.10.0 and BoTorch~0.17.2. The new oracle diagnostic uses
Python~3.14.8 with the same recorded NumPy and SciPy versions. The
\texttt{GaussianProcessRegressor} kernel is initialized as
\(\operatorname{ConstantKernel}(1,[10^{-3},10^3])\) times an ARD
Mat\'ern-\(5/2\) kernel with all seven initial length scales equal to one and
bounds \([10^{-3},10^3]\), plus
\(\operatorname{WhiteKernel}(10^{-6},[10^{-10},10^{-1}])\).
Targets are standardized externally, so \texttt{normalize\_y=False}; the
regressor uses \texttt{alpha=0}, scikit-learn's L-BFGS-B optimizer, eight
optimizer restarts, and the split seed as \texttt{random\_state}. Thus the
standardized GP has a zero prior mean rather than an independently fitted
constant-mean parameter.

The seven length scales \(\ell_j\), signal variance \(\sigma_f^2\), and white
term \(\sigma_n^2\) are estimated by maximizing the log marginal likelihood.
Because the FEM responses are deterministic, the white term is interpreted as
a numerical nugget that absorbs solver, interpolation, and model-fit
discrepancies rather than as physical measurement noise. Across the five
200-sample fits, its optimized value ranged from
\(1.0\times10^{-10}\) to \(3.04\times10^{-5}\) on the standardized target
scale. Two convergence warnings were recorded across these fits; some
smaller-sample learning-curve fits also reached a kernel bound.

For a query \(\mathbf{u}_*\), standard GP conditioning provides predictive
mean \(\mu_*(\mathbf{u}_*)\) and variance \(\sigma_*^2(\mathbf{u}_*)\):
\begin{equation}
\mu_*
=
m_*+\mathbf{k}_*^{\mathsf T}\mathbf{K}^{-1}
(\widetilde{\mathbf{y}}-\mathbf{m}),
\qquad
\sigma_*^2
=
k_{**}-\mathbf{k}_*^{\mathsf T}\mathbf{K}^{-1}\mathbf{k}_* .
\label{eq:compact_gp_posterior}
\end{equation}
Both quantities are transformed back to the physical response scale. The
predictive mean supplies the point estimate. Held-out calibration uses
scikit-learn's full predictive standard deviation, which includes the fitted
\texttt{WhiteKernel} contribution at a test point. For deterministic
acquisition, sequential search uses latent-function
variance: the training covariance retains the fitted nugget, while the
test-point diagonal and cross-covariance use only the signal kernel.

\subsection{Surrogate Validation}
Five train--validation splits use 50 held-out observations and nested training
subsets of 25--200 observations. No validation observation enters
preprocessing or fitting. We report $R^2$, RMSE, MAE, and empirical coverage
of predictive intervals. Residual, calibration, and ARD diagnostics
are included in the reproducibility archive; they are not evidence of global optimization
or calibrated uncertainty everywhere in the feasible domain.

\subsection{Surrogate Accuracy and Calibration}
\label{sec:sample_complexity_results}
At 200 training observations, the five-split mean $R^2$ is
$0.9996\pm0.0002$, RMSE $3.29\pm0.84\ \mu$T, and MAE
$2.09\pm0.29\ \mu$T. Mean one- and two-standard-deviation coverage is
0.736 and 0.908, respectively. At 25 observations, mean RMSE is
$20.32\pm2.92\ \mu$T and two-standard-deviation coverage is 0.776.
Thus prediction accuracy improves with data, but the uncertainty is not
uniformly calibrated. These held-out results describe the sampled coil
distribution and do not certify boundary extrapolation or online optima.
The smooth fitted response motivates the GP-mean control, whose scope is
tested separately on the thermal problem.

\FloatBarrier
\section{Numerical verification of selected designs}
\subsection{Numerical Verification of Selected Designs}
\label{sec:v3_validation_protocol}
The strongest observed EI-FD, GP-mean, and pool-EI designs are
selected separately for fixed current density and fixed total current. Each
is freshly re-solved with COMSOL automatic mesh levels 3, 2, and 1 at the
200-mm outer domain, at 300 and 400 mm with mesh level 3, and with the combined
finest-mesh, 400-mm setting. The fixed-current cases retain the 25-A override
during all re-solves. A hash comparison verifies that these in-memory changes
do not alter the archived source model. These are numerical checks at the
reported designs, not re-optimizations or physical measurements.
The strongest observed design of each automatic-gradient EI and LogEI control
receives the same six checks after its ten optimization runs are complete.

For conduction, each seed's terminal winner is re-evaluated at maximum mesh
spacings of 0.625 and 0.3125 mm. The same designs are used on all meshes, so
the comparison directly tests the numerical persistence of the response
differences. The shared, independent FEM implementation also permits
inspection of the temperature field, material layout, energy balance, and
algebraic residual at a representative winner.

\subsection{Numerical Checks of Selected Coil Designs}
\label{sec:v3_winner_results}
Table~\ref{tab:v3_winners} reports eight policy/excitation entries (seven distinct geometries) at production resolution and
at the combined finest mesh and largest domain. Each row records six fresh
stationary solves; the table displays the joint endpoint of the mesh--domain
study rather than combining two separately measured error estimates.
\begin{table}[!tb]\centering\small
\caption{Fresh numerical checks of the strongest observed coil designs for the listed policies.
Responses are in mT; ``fine/large'' uses mesh level 1 and a 400-mm domain.
Fixed-total-current rows come from the ten-seed study in Appendix~\ref{sec:fixed_current_protocol};
fixed-current-density EI-FD, Pool EI and Mean-FD rows are selected from the
twenty-seed study in Appendix~\ref{app:extended}, while EI-AD and LogEI-AD
come from the ten-seed matched controls.}
\label{tab:v3_winners}
\begin{tabular}{llrrr}\toprule
Excitation & Policy & Seed & Production & Fine/large\\\midrule
Fixed total current & EI-FD & 51 & 1.07453 & 1.07584 \\
Fixed total current & Pool EI & 45 & 1.06442 & 1.05951 \\
Fixed total current & Mean-FD & 49 & 1.07465 & 1.07578 \\
Fixed current density & EI-AD & 51 & 0.97708 & 0.97926 \\
Fixed current density & EI-FD & 51 & 0.97708 & 0.97926 \\
Fixed current density & Pool EI & 56 & 0.94276 & 0.94602 \\
Fixed current density & LogEI-AD & 45 & 0.97841 & 0.98099 \\
Fixed current density & Mean-FD & 52 & 0.97996 & 0.97806 \\
\bottomrule\end{tabular}\end{table}
The EI-AD and EI-FD seed-51 entries are the same
geometry, found at the first acquisition by both policies. Their repeated
re-solves check repeatability, not independent design confirmation.
Both finite-difference policies retain their advantage over the pool winner in
both excitation formulations at the joint fine/large setting. The two
finite-difference policies can exchange their nominal ordering as numerical
resolution changes; the larger distinction is between continuous feasible
search and finite-pool search. The source-model SHA-256 is identical before
and after the 48 verification solves. Intermediate mesh settings, domain sizes, and raw
responses are retained in the experiment archive.

\subsection{Selected Geometry}
\label{sec:optimized_design_results}
The strongest finite-difference EI coil design (seed 51) has
$(c_1,r_1,t_1,l_1,l_2,w_1,h_1)\approx(0.5,10.42,10,7,20,5,5)$ mm,
with core mass at 200 g and zero axial-fit margin. Its production response
is 0.97708 mT and the joint fine/large value is 0.97926 mT
(Table~\ref{tab:v3_winners}). The strongest GP-mean design (seed 52) changes
from 0.97996 to 0.97806 mT under the same joint check. These changes reinforce
the distinction between a large pool-search deficit and small unresolved
differences among continuous policies. None of the selected designs is
certified globally optimal.

\FloatBarrier
\section{Extended coil comparison}
\label{app:extended}
\subsection{COMSOL-Embedded Reference Optimizers}

BOBYQA, COBYLA, EGO, and Nelder--Mead were run through the COMSOL
optimization interface using the same parameterized FEM model, variable bounds,
constitutive representation, active coil excitation, and analytical constraints. Each method
directly evaluated \(B_{\mathrm{ROI}}\); no external GP prediction replaced its
FEM objective.

For each seed, comparison accounting starts from the same 25 shared initial
observations used by EI-BO. The embedded optimizer itself does not fit those
observations; it starts from a common interior feasible design selected
deterministically from that seed's shared set. Its best-so-far continuation is
combined with the shared initial incumbent. Thus the protocol shares the
initial incumbent and accounted FEM history, but it does not provide equal use
of the 25 observations: the surrogate methods fit all of them, whereas an
embedded method starts from the best member of the set that lies at least
\(0.1\) inside every normalized bound. This difference is inherent to the
compared workflows.
Each continuation has an accounted budget of
\begin{equation}
N_{\mathrm{cont}}^{\max}=50,\qquad
N_{\mathrm{total}}^{\max}=25+50=75.
\label{eq:compact_fem_budget}
\end{equation}
Native solver stopping conditions remained active, so an embedded continuation
could terminate before using all 50 calls. Trial points that violate an
analytic constraint are still solved by the embedded optimizers and therefore
consume accounted FEM observations, but only feasible designs can become the
incumbent. All methods were repeated for the 20 seeds 42--61, permitting
paired comparisons at common accounted counts.
\subsection{Comparison, Cost Accounting, and Rerun}

Python controls normalization, feasibility checks, GP fitting, EI evaluation,
and data logging; MPh transfers physical parameters to COMSOL, which rebuilds
the geometry and mesh, solves the axisymmetric model, and returns the
axisymmetric surface average of \texttt{mf.normB}. The 25 initial FEM
observations and up to 50 retained continuation observations are included in
every method's accounted cumulative budget.

An \emph{accounted FEM observation} is either one shared initial observation or
one retained finite continuation objective within the configured cap. A raw
solver execution, a finite objective return, and a unique feasible geometry are
stored separately. COMSOL can emit a terminal or restart row after the nominal
cap; for example, EGO seed 43 contains 51 finite continuation rows and hence 76
raw observations, while its accounted comparison budget remains 75. Across the
EGO runs of seeds 42--46 the mean raw count is 75.2. These extra rows are implementation
overhead and are not silently described as unique-design budget slots.

The primary horizontal axis is therefore the number of accounted FEM
observations, not every low-level solver execution.
Performance is reported through best-so-far trajectories and the best
feasible response available after 26, 30, 40, 50, and 75 accounted
observations, summarized by the mean and sample standard deviation over the
20 paired seeds. A natively early-stopped run retains its terminal incumbent
at later budgets, because that design remains the best one available to the
user.

In the seven-policy coil experiment, EI-FD is compared
with the other six methods on the 20
seed-matched differences at 30, 50, and 75 observations. We report the mean
difference with its 95\% Student-\(t\) interval, the number of seeds won, the
paired effect size \(d_z\), and a two-sided Wilcoxon signed-rank test with
Holm correction over the six comparisons at each budget. The
ten-seed continuous-control comparisons report unadjusted two-sided
Wilcoxon values and are not included in that correction family.

End-to-end elapsed time is reported separately. It includes initial-data
generation, Python--MPh communication, GP refitting, candidate selection, and
COMSOL execution for EI, whereas embedded times include the corresponding
internal COMSOL runs. These measurements describe implementation cost but do
not provide a controlled wall-clock comparison.

The best observed EI design is transferred through the same
Python--MPh--COMSOL path and solved again. Agreement checks parameter
reconstruction and software-path repeatability only; it is not independent
physical validation. The final GP prediction and standardized residual at that
design are reported as local surrogate diagnostics.

\subsection{Extended Coil Comparison with Embedded Optimizers}
\label{sec:fixed_budget_results}

\begin{figure}[!tb]
    \centering
    \includegraphics[width=0.86\textwidth]{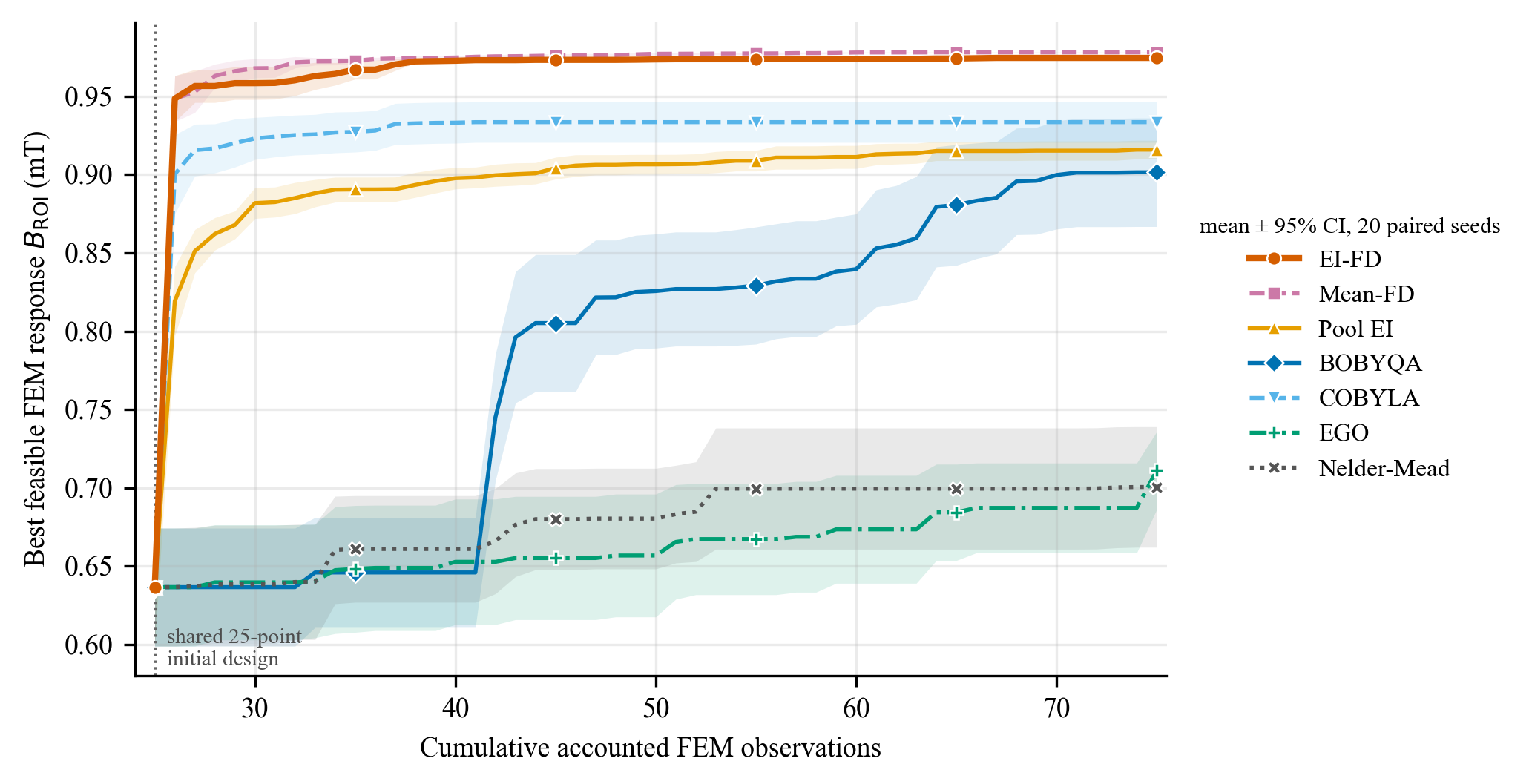}
    \caption{Best feasible FEM-observed response versus cumulative accounted
    FEM observations over 20 paired seeds. Every trajectory starts from the
    same seed-specific 25-point initial design. Bands are pointwise 95\%
    Student-\(t\) intervals.}
    \label{fig:best_fem_response_vs_evaluations}
\end{figure}

Each run begins with 25 shared initial observations and permits up to 50
method-specific continuation evaluations. The seven policies were run for the
20 seeds 42--61. Figure~\ref{fig:best_fem_response_vs_evaluations} shows the
best-so-far trajectories, Table~\ref{tab:fixed_budget_optimization_results}
the budget checkpoints, and Table~\ref{tab:budget_checkpoints} the paired
comparisons. Bands in the figure are pointwise two-sided 95\% Student-\(t\)
intervals over the 20 seeds, not simultaneous confidence bands.

\emph{Continuous refinement improves on the pool and embedded references.} From a common mean
initial incumbent of \(0.637\) mT, EI-FD reaches
\(0.949\pm0.031\) mT after a single new FEM solve and
\(0.975\pm0.002\) mT at the end of the budget. The corresponding
terminal values are \(0.916\pm0.013\) mT for pool EI,
\(0.934\pm0.028\) mT for COBYLA, \(0.901\pm0.074\) mT for BOBYQA,
\(0.711\pm0.053\) mT for EGO, and \(0.701\pm0.082\) mT for Nelder--Mead.
EI-FD has the higher terminal response on all 20 seeds
against each of these five methods, with Holm-adjusted \(p<10^{-3}\) in every
case (Table~\ref{tab:budget_checkpoints}). All 20 of its runs exceed
\(0.95\) mT, with a median of one new solve; no pool EI run and only seven
BOBYQA and seven COBYLA runs reach that level within 75 observations. Its
worst seed (\(0.973\) mT) is within \(0.001\) mT of the best BOBYQA
seed (\(0.974\) mT) and above the best pool EI
seed (\(0.943\) mT).

\emph{Continuous refinement reduces variation across seeds.} The terminal sample SD of
EI-FD is \(0.002\) mT, against
\(0.013\) mT for pool EI, \(0.028\) mT for
COBYLA, and \(0.074\) mT for BOBYQA. With 20 seeds the
derivative-free references are visibly start-dependent: BOBYQA terminates
between \(0.676\) and \(0.974\) mT,
and COBYLA stops after a mean of 39.1
accounted observations at a design that it does not subsequently improve.

\emph{Analytic screening avoids infeasible FEM evaluations.} The
COMSOL-embedded optimizers solve trial designs that violate the analytic
constraints: on average 22.3
of BOBYQA's 50 continuation solves, 17.6
of EGO's, and 11.6 of COBYLA's
are infeasible and cannot become the incumbent. The surrogate-based methods
screen or enforce the constraints before the FEM call and spend none.

\emph{The coil gain does not require uncertainty-aware acquisition.}
Pool and EI-FD share the acquisition definition and initial
data but differ in the refinement step, accepted in 95\% of iterations.
Their terminal means differ by 0.059 mT. The GP-mean control reaches
$0.978\pm0.001$ mT and nominally exceeds EI, although its advantage is
comparable to mesh sensitivity. This is evidence that continuous exploitation
is sufficient on this coil response, not that uncertainty is unimportant
in general. The thermal seed-51 outlier demonstrates the limitation of
extending that conclusion across problems.

\emph{The observed designs approach intersecting active constraints.}
Every terminal design of the two finite-difference policies has \(t_1\),
\(w_1\), and \(h_1\) on their upper bounds and \(l_1\) on the axial-fit
constraint, and the core mass lies between 191 and 200 g, on the 200-g limit
in all GP-mean runs. EI-FD additionally places \(c_1\) on its
lower bound and \(l_2\) on its upper bound. Between five and seven constraints
are therefore active, including the nonlinear mass limit. Pool EI moves
toward the same vertex, with terminal \(w_1\) and \(h_1\) at 95--99\% of
their ranges, but does not reach the same boundary intersection in the reported runs.
This is consistent with the geometric mechanism, without applying the
uniform-box formula literally to nonlinear feasible-domain LHS sampling.

\begin{table}[!tb]
    \centering
    \caption{Best feasible FEM-observed response (mT) after \(n\) accounted FEM
    observations, mean \(\pm\) sample SD over 20 paired seeds. All methods share
    the seed-specific 25-point initial design (mean incumbent \(0.637\) mT).
    ``Reached'' counts runs attaining \(0.95\) mT within the budget;
    ``Infeasible'' is the mean number of continuation solves spent on designs
    that violate an analytic constraint.}
    \label{tab:fixed_budget_optimization_results}
    \small
    \resizebox{\textwidth}{!}{%
\begin{tabular}{lccccccc}
    \toprule
    Method & \(n=26\) & \(n=30\) & \(n=40\) & \(n=50\) & \(n=75\) & Reached & Infeasible \\
    \midrule
    EI-FD & \(0.949\pm0.031\) & \(0.958\pm0.022\) & \(0.973\pm0.002\) & \(0.974\pm0.002\) & \(0.975\pm0.002\) & 20/20 & 0.0 \\
    Mean-FD & \(0.948\pm0.031\) & \(0.968\pm0.013\) & \(0.975\pm0.002\) & \(0.977\pm0.002\) & \(0.978\pm0.001\) & 20/20 & 0.0 \\
    Pool EI & \(0.819\pm0.047\) & \(0.882\pm0.021\) & \(0.898\pm0.015\) & \(0.907\pm0.013\) & \(0.916\pm0.013\) & 0/20 & 0.0 \\
    COBYLA & \(0.900\pm0.054\) & \(0.923\pm0.029\) & \(0.933\pm0.028\) & \(0.934\pm0.028\) & \(0.934\pm0.028\) & 7/20 & 11.6 \\
    BOBYQA & \(0.637\pm0.080\) & \(0.637\pm0.080\) & \(0.646\pm0.075\) & \(0.826\pm0.078\) & \(0.901\pm0.074\) & 7/20 & 22.3 \\
    EGO & \(0.637\pm0.080\) & \(0.640\pm0.079\) & \(0.653\pm0.086\) & \(0.657\pm0.084\) & \(0.711\pm0.053\) & 0/20 & 17.6 \\
    Nelder--Mead & \(0.637\pm0.080\) & \(0.638\pm0.081\) & \(0.661\pm0.073\) & \(0.680\pm0.069\) & \(0.701\pm0.082\) & 0/20 & 0.0 \\
    \bottomrule
    \end{tabular}%
}
\end{table}

\begin{table}[!tb]
    \centering
    \caption{Paired differences, EI-FD minus the listed method
    (mT), over 20 seeds. Positive values favor EI-FD. ``Wins''
    counts seeds with a strictly positive difference; \(p\) is the Holm-adjusted
    two-sided Wilcoxon signed-rank value within each budget.}
    \label{tab:budget_checkpoints}
    \small
    \begin{tabular}{clccccc}
    \toprule
    \(n\) & Compared method & Mean difference & 95\% interval & Wins & \(d_z\) & \(p\) \\
    \midrule
    30 & Mean-FD & \(-0.009\) & \([-0.019,\,+0.000]\) & 5/20 & \(-0.46\) & \(0.055\) \\
    30 & Pool EI & \(+0.077\) & \([+0.064,\,+0.090]\) & 20/20 & \(+2.75\) & \(<10^{-3}\) \\
    30 & COBYLA & \(+0.035\) & \([+0.018,\,+0.053]\) & 18/20 & \(+0.94\) & \(<10^{-3}\) \\
    30 & BOBYQA & \(+0.322\) & \([+0.284,\,+0.360]\) & 20/20 & \(+3.95\) & \(<10^{-3}\) \\
    30 & EGO & \(+0.319\) & \([+0.281,\,+0.356]\) & 20/20 & \(+3.99\) & \(<10^{-3}\) \\
    30 & Nelder--Mead & \(+0.320\) & \([+0.282,\,+0.358]\) & 20/20 & \(+3.93\) & \(<10^{-3}\) \\
    \midrule
    50 & Mean-FD & \(-0.004\) & \([-0.005,\,-0.003]\) & 1/20 & \(-1.70\) & \(<10^{-3}\) \\
    50 & Pool EI & \(+0.067\) & \([+0.061,\,+0.073]\) & 20/20 & \(+4.86\) & \(<10^{-3}\) \\
    50 & COBYLA & \(+0.040\) & \([+0.027,\,+0.053]\) & 20/20 & \(+1.43\) & \(<10^{-3}\) \\
    50 & BOBYQA & \(+0.148\) & \([+0.111,\,+0.185]\) & 20/20 & \(+1.89\) & \(<10^{-3}\) \\
    50 & EGO & \(+0.317\) & \([+0.278,\,+0.355]\) & 20/20 & \(+3.82\) & \(<10^{-3}\) \\
    50 & Nelder--Mead & \(+0.293\) & \([+0.261,\,+0.325]\) & 20/20 & \(+4.30\) & \(<10^{-3}\) \\
    \midrule
    75 & Mean-FD & \(-0.004\) & \([-0.004,\,-0.003]\) & 0/20 & \(-2.01\) & \(<10^{-3}\) \\
    75 & Pool EI & \(+0.059\) & \([+0.053,\,+0.064]\) & 20/20 & \(+4.69\) & \(<10^{-3}\) \\
    75 & COBYLA & \(+0.041\) & \([+0.028,\,+0.054]\) & 20/20 & \(+1.47\) & \(<10^{-3}\) \\
    75 & BOBYQA & \(+0.073\) & \([+0.038,\,+0.108]\) & 20/20 & \(+0.98\) & \(<10^{-3}\) \\
    75 & EGO & \(+0.263\) & \([+0.239,\,+0.288]\) & 20/20 & \(+5.01\) & \(<10^{-3}\) \\
    75 & Nelder--Mead & \(+0.274\) & \([+0.236,\,+0.312]\) & 20/20 & \(+3.36\) & \(<10^{-3}\) \\
    \bottomrule
    \end{tabular}
\end{table}

\subsection{Evaluation Cost}
\label{sec:evaluation_efficiency_results}

End-to-end elapsed times were recorded for the
five-seed timing subset (seeds 42--46) of pool EI and the embedded optimizers. The FEM
solves of this benchmark take about two seconds, so per-iteration overhead is
visible: mean end-to-end time was 178.3 s for pool EI, compared with
116.4 s for BOBYQA, 91.6 s for COBYLA, 109.6 s for EGO, and 122.0 s for
Nelder--Mead. The finite-difference refinement adds approximately \(0.2\) s of
acquisition time per iteration to pool EI, about 10 s over a 50-acquisition
run, and no FEM call.

The relevant comparison is therefore cost to a given design quality. Reaching
\(0.95\) mT takes EI-FD a median of one new FEM solve,
whereas BOBYQA and COBYLA each reach it in seven of 20 runs; EGO and
Nelder--Mead do not reach it within the accounted budget. Longer FEM solves
could increase the value of avoiding evaluations, but the experiments do
not measure runtime savings on industrial-scale models.

\FloatBarrier
\section{Fixed-total-current coil variant}
\subsection{Fixed-current protocol}
\label{sec:fixed_current_protocol}

The production objective excites the coil at fixed current density, so the
total current grows with the coil cross-section. To verify that the
conclusions do not depend on this choice, the complete design space is
re-optimized with the coil current overridden to a geometry-independent
\(I_{\mathrm{coil}}=25\) A, the largest production value. The override is
applied in memory after the archived model contract has been validated, and
the source model file is not modified. For seeds 42--51, the same 25 initial
geometries are re-solved under the new excitation and shared by pool EI,
EI-FD, and Mean-FD, each followed by 50
acquisitions. Neither the COMSOL-embedded optimizers nor the automatic-gradient EI and
LogEI controls are rerun in this variant; no ordering against them is inferred.

\subsection{Fixed-current results}
\label{sec:fixed_current_results}

Table~\ref{tab:fixed_current_results} reports the variant of
Appendix~\ref{sec:fixed_current_protocol}, in which every geometry carries
25 A. Removing the coupling between coil area and current changes the
optimum qualitatively. The best designs now use the smallest admissible coil
(\(h_1=0.5\) mm, \(w_1\) between \(0.5\) and \(1.1\) mm) and the shortest
lower core section (\(l_1=2.5\) mm, where the axial-fit constraint is again
active), while \(t_1\) remains on its upper bound and the core mass on its
200-g limit. The active set therefore differs from that of the
fixed-current-density problem, in which \(w_1\) and \(h_1\) were on their
upper bounds.

Among the three policies tested in this variant, the ordering is unchanged. EI-FD reaches
\(1.074\pm0.001\) mT after 75 observations against \(1.058\pm0.005\) mT for
pool EI, a paired difference of \(0.016\) mT (95\% interval
\([0.013,0.019]\) mT) that is positive on all 10 seeds (Wilcoxon
\(p=0.002\)). The pool designs again stop short of the bounds, with no
coordinate exactly on a box bound in any run. The margin is smaller than in
the production problem, \(1.5\%\) instead of \(6\%\), and at 30 observations the difference is not yet
resolved (\(0.012\) mT, interval \([-0.002,0.026]\) mT). The GP-mean control
reaches \(1.075\) mT on every seed and is again indistinguishable from EI
within the numerical accuracy of the model.

\begin{table}[!tb]
    \centering
    \caption{Fixed-total-current re-optimization (25 A): best feasible
    FEM-observed response (mT), mean \(\pm\) sample SD over 10 paired seeds.
    The shared initial designs have a mean incumbent of \(0.936\) mT.}
    \label{tab:fixed_current_results}
    \small
    \begin{tabular}{lccc}
    \toprule
    Method & \(n=30\) & \(n=50\) & \(n=75\) \\
    \midrule
    EI-FD & \(1.058\pm0.019\) & \(1.074\pm0.002\) & \(1.074\pm0.001\) \\
    Mean-FD & \(1.059\pm0.013\) & \(1.075\pm0.000\) & \(1.075\pm0.000\) \\
    Pool EI & \(1.046\pm0.012\) & \(1.054\pm0.006\) & \(1.058\pm0.005\) \\
    \bottomrule
    \end{tabular}
\end{table}

\FloatBarrier
\section{Supplementary Numerical Records}
\label{app:numerical-records}
The records described here are retained in the separate reproducibility
archive; they do not accompany this arXiv version. Its README maps files to
experiments and gives execution commands.
\texttt{matched\_terminal\_results.csv} contains the 100 original terminal
outcomes; \texttt{best} is coil response in tesla or negative thermal
resistance in K/W. Identifiers \texttt{pool}, \texttt{ei\_cas},
\texttt{mean\_cas}, \texttt{ei\_ad}, and \texttt{logei\_ad} denote Pool EI,
EI-FD, Mean-FD, EI-AD, and LogEI-AD.

Files prefixed \texttt{extended\_coil}, \texttt{fixed\_current},
\texttt{oracle}, and \texttt{thermal\_boundary} record the corresponding
studies. The oracle archive includes all 3840 outcomes; the additional
thermal control includes all ten trajectories and 30 mesh checks.
\texttt{active\_constraint\_scaling.csv} records dimensionless regret and
bound gaps across eight seeds per configuration. Numerical-check files
\texttt{thermal\_mesh\_validation.csv} and
\texttt{coil\_winner\_validation.csv} specify units in their headers.

\end{document}